\documentclass{article}

\usepackage{arxiv}

\usepackage[utf8]{inputenc} 
\usepackage[T1]{fontenc}    
\usepackage{hyperref}       
\usepackage{url}            
\usepackage{booktabs}       
\usepackage{amsfonts}       
\usepackage{nicefrac}       
\usepackage{microtype}      
\usepackage{lipsum}
\usepackage{graphicx}
\usepackage{booktabs}
\graphicspath{ {./images/} }
\usepackage[normalem]{ulem}
\useunder{\uline}{\ul}{}
\usepackage{multirow}
\usepackage{threeparttable, tablefootnote}
\usepackage{caption}
\usepackage{subcaption}
\usepackage{amsmath}

\usepackage[natbib=false,style=ext-numeric,sorting=none]{biblatex}
\bibliography{main.bib}

\usepackage{color}

\hypersetup{
  colorlinks   = true, 
  urlcolor     = blue, 
  linkcolor    = blue, 
  citecolor   = blue 
}

\title{domain adaptation for Arabic machine translation: the case of financial texts}

\author{
 Emad A. Alghamdi\\
  King Abdulaziz University \\
  \texttt{eaalghamdi@kau.edu.sa} \\
   ASAS AI Lab \\
   \And
 Jezia Zakraoui  \\
  \texttt{j.zakraoui@gmail.com} \\
    ASAS AI Lab \\
  \And
 Fares A. Abanmy\\
  \texttt{abanmyfares@gmail.com} \\
    ASAS AI Lab \\
}

\begin{document}
\maketitle

\begin{abstract}
Neural machine translation (NMT) has shown impressive performance when trained on large-scale corpora. However, generic NMT systems have demonstrated poor performance on out-of-domain translation. To mitigate this issue, several domain adaptation methods have recently been proposed which often lead to better translation quality than genetic NMT systems. While there has been some continuous progress in NMT for English and other European languages, domain adaption in Arabic has received little attention in the literature. The current study, therefore, aims to explore the effectiveness of domain-specific adaptation for Arabic MT (AMT), in yet unexplored domain, financial news articles. To this end, we developed carefully a parallel corpus for Arabic-English (AR-EN) translation in the financial domain for benchmarking different domain adaptation methods. We then fine-tuned several pre-trained NMT and  
Large Language models including 
ChatGPT-3.5 Turbo on our dataset. The results showed that the fine-tuning is successful using just a few well-aligned in-domain AR-EN segments. The quality of ChatGPT translation was superior than other models based on automatic and human evaluations. To the best of our knowledge, this is the first work on fine-tuning ChatGPT towards financial domain transfer learning. To contribute to research in domain translation, we made our datasets and fine-tuned models available at \href{https://huggingface.co/asas-ai/}{https://huggingface.co/asas-ai/}. 





\end{abstract}
\keywords{Machine Translation \and Arabic MT \and Domain Adaptation, \and Financial Domain}

\subsection{Introduction}
In the recent years, the rapid advancement of deep learning techniques and their adaptation in machine translation has made a great stride in many translation tasks. Neural Machine Translation (NMT) systems, trained on a large-scale corpora, have demonstrated impressive performance in translating generic language. However, NMT models tend to perform poorly on out-of-domain data \cite{koehn2017six}, especially if the target domain has a distinctive style and vocabulary \cite{daume2011domain}. A NMT model trained on exclusively medical texts is unlikely to achieve accurate performance on financial or news data. To address this problem, researchers have proposed different domain adaptation approaches and techniques which seem to improve the quality of NMT systems on out-of-domain data \cite{saunders2022domain, chu2020survey, moslem2023adaptive, }. 

While there are many MT models, systems and tools for translating Arabic texts in the literature; however, the quality of the translation is poor, especially for out-of-domain texts \cite{popel2020transforming}. A key technical challenge related to AMT arises from the lack of available bilingual datasets for out-of-domain texts that can be used as standard benchmark to conduct unified experiments. In fact, researchers tend to collect datasets according to their specific domains and try to resolve the linguistic issues for Arabic, based on custom datasets such as in the domain of news \cite{hatem2010syntactic}, \cite{almahasees2018assessment} ignoring hereby many other domains. Other technical issues such as out-of-vocabulary (OOV) and very long sentences also make MT more challenging \cite{koehn2017six}. To address these challenges, researchers have proposed different techniques, including for example, BPE \cite{sennrich2015neural}, character-level BPE variant \cite{costa2016character}, hybrid techniques \cite{luong2016achieving}, and mixed fine-tuning \cite{popel2020transforming}. However, domain robustness remains an unsolved problem and there is a need for further research in this area \cite{muller2019domain}. This is specially true for Arabic language. Existing domain adaptation research has only focused on news \cite{oudah2019impact} and medical \cite{moslem2022domain} domains, no prior study, to the best of our knowledge, has been conducted on financial domain. 

To alleviate the issue with translation mismatch related to out-of-domain texts, the authors in \cite{oudah2019impact} studied the performance of NMT systems under morphology-based and frequency-based tokenization schemes and BPE on in-domain data. They evaluated their best performing models on out-of-domain data yielding significant improvements of 37.96\% in BLEU score \cite{papineni2002bleu}. The latter \cite{moslem2022domain} proposed a method for domain-specific data augmentation for MT to tackle the issue with a small bilingual dataset. They employed mixed fine-tuning to train models that significantly improve translation of in-domain texts. Their method achieved improvements of approximately 5-6 BLEU and 2-3 BLEU, respectively, on the Arabic-to-English and English-to-Arabic language pairs.

While a lot of research in domain adaptation in MT for other language pairs like \cite{popel2020transforming}, \cite{bapna2019simple} exist which focus on synthetic data generation and multiple other techniques like checkpoint averaging \cite{popel2020transforming}, only one work \cite{moslem2022domain} investigated the same for AMT, but only for a medical domain. This research aims to fill the gap in creating different MT settings and investigate in the domain of financial texts and potentially to extend to other domains.

Our contributions are the following:
\begin{itemize}
    \item We introduce the first AR-EN parallel corpus in the financial domain. 
    \item We compare the effectiveness of different adaption methods and data augmentation approaches for limited domain data. 
    \item We fine-tuned several models and made them publicly available to the research community.
    \item Our work is the first to fine-tune GPT3.5 model and evaluate its capability for domain adaption. 
\end{itemize}


\section{Background}

\subsection{Neural machine translation}
NMT models based on deep neural networks (DNN) have been proposed in early NMT research \cite{yang2020survey}. A DNN based NMT model employs a neural network system to perform the required machine translation tasks using an encoder-decoder network \cite{sutskever2014sequence}. The encoder neural network inputs and encodes a source language sentence into a fixed-length vector in each hidden state. Then, given the final hidden state of the encoder, the decoder does the reverse work by transforming the hidden state vector to the target sentence word by word. A translation probability of a source sentence is modeled into the target sentence. Given a source sentences 
$S$= $\bigl\{s_1, s_2, ..s_n \bigl\}$
and a target sentence 
$T$= $\bigl\{ t_1, t_2, ..t_n \bigl\}$ 
, the encoder encodes all the words from the source sentence $S$ 
into a set of hidden states $\bigl( h_1, h_2, ..h_n \bigl) $ 
and passes the fixed-size vector $v$, which represents the source sentence, to the decoder. The translation probability with a single neural network is given by following formula \cite{johnson2017google}:

\begin{equation}
    P(S)= \prod_{i=1}^{n} P(t_{<i},S)
\end{equation}

where $t<i$ stands for the sequence preceding the $ith$ target word.
Hence each predicted word $t_{i}$ is based on the previously predicted word $t_{i-1}$ and the previous hidden states $h_{i-1}$. However, when the sentences become long the performance deteriorates. This limitation is due to the limited feature representation ability in a fixed-length vector \cite{yang2020survey}. To overcome this issue and to provide additional word alignment information in translating long sentences, Bahdanau et al. \cite{bahdanau2014neural} introduced the idea of the attention mechanism. Concretely, attention mechanism is an intermediate component between encoder and decoder, which can help to determine the word alignment dynamically. The decoder pays attention to input or to any part of the input sentence. Attention is calculated using each encoder output and the current hidden state, resulting in a vector of the same size as the input sequences using score functions \cite{bahdanau2014neural}. There are mainly three different architectures for constructing NMT, namely Recurrent neural network (RNN), Convolution neural network (CNN), and Self-attention-based Transformer. 

The use of RNN-based models has demonstrated good quality translation results. This type of network is composed of encoder and decoder with similar working of sequence-to-sequence learning. Multiple variant of RNN architectures include i.e., LSTM \cite{zhou2016deep}, BiLSTM \cite{bahdanau2014neural} and GRU \cite{ataman2019latent}.
   
The second approach of developing NMT systems based in convolution neural network (CNN) architecture. Work using CNN has generally reported good results, specially for word-based MT \cite{meng2015encoding}. This work applied a convolution layer on the bottom of the recurrent layer which hinders the performance. The bottleneck was handled by implementing the fully convolutional model as suggested by \cite{gehring2016convolutional}. The performance and accuracy were improved with a number of models; word-based \cite{kalchbrenner2016neural}, character-based \cite{costa2016character}, and recently with attention \cite{gehring2017convolutional}.

Recently, the use of transformers has resulted in well-performing machine translation systems. This type of it is a sequence-to-sequence model \cite{vaswani2017attention}, which consists of a stack of layers. Each layer first utilizes the self-attention to extract information from the whole sentence, then follows a point-wise feed-forward network to provide non-linearity. The novel idea of self-attention is to extend the mechanism to the processing of input sequences and output sentences as well. In general form, the Transformer attention function uses three vectors: queries(Q), keys (K) and values (V).

The output is a weighted sum of values, where weights are computed by a similarity score between $n$ query vectors and $m$ keys \cite{vaswani2017attention}. The attention is defined as follows: 

\begin{equation}
    Attention(Q,K,V)=softmax(score(Q,K))V
\end{equation}

where score Q,K is an n×m matrix of similarity scores. A straightforward choice for scoreQ,K proposed by Luong et al. \cite{luong2015effective} is the dot product i.e. score(Q,K) =QK. The softmax function normalizes over the columns of that matrix so that the weights for each query vector sum up to one. There are many variants in the implementation of attention-based models which are classified into two broad categories, global and local attention discussed in detail in this survey \cite{yang2020survey}.

Current state-of-the-art NMT models \cite{stahlberg2020neural} rely on the Transformer model \cite{vaswani2017attention} and multiple attention mechanism \cite{bahdanau2014neural}. However, the transformer-based language models such as Bidirectional Encoder Representation from Transformers (BERT) \cite{devlin2018bert} expands the function of attention to encompass the main task. It uses self-attention, which is applied to two states within the same sequence, as the foundation for sequence representations rather than an RNN. For Arabic language, two transformer-based language models have been developed so far; notably AraBERT \cite{antoun2020arabert} and GigaBERT \cite{lan2020gigabert}. Both models aim at solving a masked language-modelling task in order to correctly predict a masked word from its context. Besides, these models aim at resolving a next sentence prediction task especially to decide whether two sentences are consecutive or not.

\subsection{Domain-specific MT}
Domain translation is a challenging task due to the fact that language varies across different domains, genres, and styles. For example, texts in a financial domain often contain specific terminologies and jargon that may not be extensively used in legal or health domains. Therefore, researchers have proposed different methods to improve the quality of translations in domains such as medical and biomedical \cite{moslem2022domain, abdul2019exploring, liu2021paramed}, legal \cite{martinez2020customized}, and financial texts \cite{laubli2019post}. Several domain adaptation approaches have been proposed \cite[for more comprehensive survey see][] {saunders2022domain}. Domain adaptation methods can intervene in various stages of NMT system design, training and use and can be classified into three main categories: data centeric methods, architecture-centric adaptation methods, and inference schemes for adaptation
\cite{saunders2022domain}. 
In data centeric methods, the objective is to select or generate appropriate in-domain data. A large generic monolingual data can be filtered to select domain-representative dataset based on some unique characteristics of the target domain. However, selecting a small in-domain dataset may be more domain relevant, but the impact of any deviation from the target domain will be magnified \cite{grangier2021trade}. Another approach is to construct partially synthetic bilingual training corpora by forward- or back translation. \cite{poncelas1804investigating} observed that models trained exclusively on back translations can perform similarly to models trained on natural data. Recently, the use of pre-trained large language models (LLMs) to generate large amounts of synthetic data at very low cost has emerged to be an effective approach \cite{moslem2022domain}.    

Architecture-centric adaptation typically involves adding trainable parameters to pre-trained models to avoid train models from scratch. A common approach is to fine-tune an existing well-performing NMT model on small in-domain data. Extensive fine-tuning can lead to catastrophic forgetting. \cite{chu2017empirical} proposed mixed-fine tuning which involves two steps: (1) training an NMT model on out-of-domain data until convergence and then (2) fine-tuning the NMT model from step 1 on a mix of in-domain and out-of-domain data (by oversampling the in-domain data) until convergence. Mixed-fine tuning approaches can be helpful to prevent two major issues notably overlooking the specificity of each domain \cite{koehn2017six} and forgetting previously learned knowledge when exposed to the new training examples as reported in \cite{deng2020factorized}. 

Lastly, the inference schemes for adaptation develop a separate NMT model to each domain and combine them at inference time. 





\subsection{Domain-adaptation in Arabic MT}
The development of Arabic MT systems has gone through different stages, including rule-based systems \cite{bakr2008hybrid, mohamed2012transforming}, statistical MT \cite{habash2009improving}, and more recently neural MT systems \cite{nagoudi2021arat5}. \cite{zakraoui2021arabic} conducted a comprehensive survey of Arabic MT systems and the unique challenges in Arabic MT.

Arabic is one of the official six languages adopted by United Nations and it is spoken by 400 million people in the middle east, north Africa, and many other parts of the world. Arabic is a Semitic language and it is notoriously difficult for MT due to its linguistic characteristics \cite{ameur2020arabic} \cite{zakraoui2021arabic}. First, Arabic has a rich and complex morphology which is substantially different from English or other western languages \cite{habash2010introduction}. Second, Arabic has long and short vowels. While the long vowels are represented by letters, the short vowels are marked by diacritic signs placed above or below the letters. However, the use of diacritic signs is not compulsatory in Arabic and hence they are rarely used in informal writing. Therefore, it is hard to identify the correct sense of a word, especially when sufficient context is not provided. Third, variation among different Arabic dialects has always been problematic for AMT. Furthermore, the Arabic language used in social medial varies considerably from Modern Standard Arabic (MSA). These aspects of the Arabic language pose series challenges for Arabic MT. 

In addition to the aforementioned issues, there is a lack of high quality parallel corpora of sufficient size for training or fine-tuning Arabic MT for different domains. It is commonly known that NMT systems do not perform well in domain specific translation, specially in low-resource languages \cite{koehn2017six}. Addressing these challenges, some researchers have turned to domain adaption methods to develop domain-specific Arabic MT systems. For example, \cite{moslem2022domain} proposed the use of pre-trained LMs and back-translation for domain-specific data augmentation for MT.


Furthermore, current Arabic MT research has primary focused on the translation of limited domains such as news and official texts, whilst few attempts focus on domain specific translation such as medical domain \cite{ehab2019english}. Specifically, most of the used parallel data available to the researcher was limited to texts produced by international organizations, parliamentary debates \cite{abdul2019exploring}. Unfortunately, existing single-domain AMT methods do not work well for multiple domains. Thus, multi-domain NMT approaches are more in demand to tackle this limitation.

To recap, previous research has shown that domain adaptation leads to better translation quality than general NMT. Since there is relatively little work on Arabic domain adaptation, the primary objective of this research is to explore different effectiveness of domain translation methods, a yet unexplored domain, financial domain. To this end, this work aims to fine-tune several Transformer NMT models and LLM and perform cross-domain testing and evaluation to gain some insights into model robustness against domain changes.

\section{Methodology}
This section gives an overview of the methods and algorithms for AMT domain adaptation using LLMs models. First, information about the collected bilingual dataset used is given which we refer as authentic dataset, then our approach is presented, and lastly, the metrics we used for evaluation are described.

\subsection{Approach}
In this work, we investigate mainly two methods to augment our in-domain data for the domain of financial news and propose approaches to leverage pre-trained LLMs for domain-specific data generation for this MT task. Concerning domain-specific data generation, we start with synthetic data generation to augment our authentic sentences for Arabic. Then, to obtain the parallel data in English, we apply back-translation from the Arabic synthetic sentences.\\
In our case, we leverage a pipeline of different models. We start by AraGPT2 \cite{antoun-etal-2021-aragpt2} and gpt2 \cite{radford2019language} as text generation models for Arabic and English to create synthetic pairs for (AR-EN) and (EN-AR), respectively. For Arabic, we use titles only from the collected authentic dataset as text prompts to generate corresponding long-form text using AraGTP2. Then, we use a summarization model to summarize the generated bunches of texts to obtain short summaries that will serve as generated titles. In the end, we back-translate the long-form text as well as the generated summaries to serve as title and article. The same pipeline applies to English as the target language. 

Figure 1 shows the case of augmenting the authentic dataset with AR-EN pairs using this method.

\begin{figure}[!htb]
\includegraphics[scale=0.5]{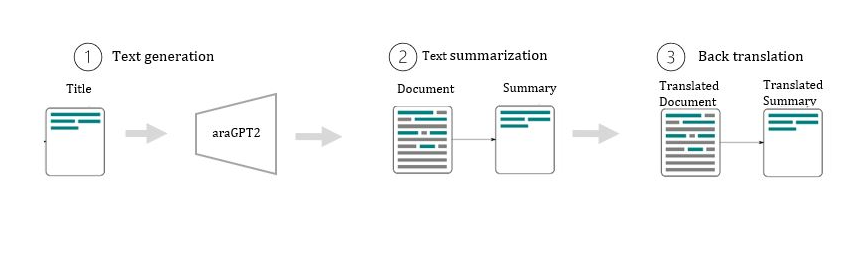}
\caption{Data augmentation pipeline}

\centering
\end{figure}

\subsubsection{Synthetic data generation}
Data augmentation has been used in domain translation due to the scarcity of domain-specific datasets that are suitable for training large models. A common approach to augment domain data is the use of back-translation when there is abundant data in the target domain \cite{poncelas1804investigating, koehn2017six}. \cite{moslem2022domain} proposed the use of state-of-the-art large language models to general unlimited new sentences in the source language and then back-translating in the target language. Recent studies explored the use of ChatGPT for generating new parallel sentences. However, in this study \cite{khondaker2023gptaraeval}, the authors showed that the performance of ChatGPT for Arabic shows inferior performance compared to the finetuned AraT5. \\




\begin{table}

\caption{Authentic dataset statistics}
\centering
\begin{tabular}{@{}lrrr@{}}
\toprule
\textbf{Source}           & \textbf{Articles} & \textbf{Titles} & \textbf{Sentences} \\ \midrule
Tadawul                   & 569               &  569            &  2544              \\
Capital Markets Authority & 2320              &  2320           &  8351             \\
Eye of Riyadh             & 891               &  891            &  1877              \\
Total                     & 3780              &  3780           &  15771             \\ \bottomrule
\end{tabular}
\label{tab:data-table}
\end{table}

\subsubsection{Back-translation}
We use a pre-trained machine translation model \cite{Tiedemann2020OPUSMTB} for back-translation. The back-translation is applied on both sides of generated summaries and titles, namely on the long-form text (which serves as an article) as well as on the summarized form (which serves as a title) into the respective target language.

\subsection{Experiment setup}

\subsubsection{Datasets}
For fine-tuning in domain-specific MT models, we collected a dataset from different online resources for the pair AR-EN. As shown in table \ref{tab:data-table} most of the data is collected from Capital Markets Authority (CMA) yielding a total of 7560 AR-EN pairs. Note that we consider titles (3780 AR-EN pairs) and articles (3780 AR-EN pairs). Additionally, we augmented our dataset with synthetic data as well as back-translated data.  This step augmented the authentic dataset by 12,318 and 12,000 AR-EN sentence pairs as synthetic and back-translated data, respectively.
\\
This table \ref{tab:datasplit-table} shows the breakdown of the segments in our dataset. We randomly sampled 1000 segments from the authentic dataset to serve as test data for all models. Additionally, we randomly sampled 1000 segments for building the development for both models notably for OPUS (bt-big) and NLLB. However, for fine-tuning chatGPT we randomly sampled 2000 pairs each for each setup.

\begin{table}
\centering
\caption{Authentic dataset split and augmented data count}
\begin{tabular}{@{}llrrr@{}}
\toprule
\textbf{Language pair} & \textbf{Type}     & \textbf{Fine-tuning} & \textbf{Dev} & \textbf{Test} \\ \midrule
AR-EN                  & Authentic         & 5560           & 1000         & 1000          \\
AR-EN                  & Synthetic         & 11318           &        1000      &       -        \\
EN-AR                  & Back-translated & 11000          &     1000         &           -    \\
                       &                   &              &              &               \\
\bottomrule
\end{tabular}

\label{tab:datasplit-table}
\end{table}

\subsubsection{NMT pre-trained models}
Our generic NMT pre-trained models use different Transformer architectures, however, we have implemented the fine-tuning objective using the huggingface NMT transformer (a sequence-to-sequence version in the Transformers library) procedure. 
For inference, we use beam size 4 and batch size 16, on a GPU T4-15GB (Google Colab). Further, we use chatGPT as a baseline with zero-shot learning.
\subsubsubsection{OPUS (bt-big):}
We use OPUS \cite{tiedemann-2020-tatoeba} models from the Tatoeba-Challenge, specifically
the models augmented with back-translated data of Wikimedia content and trained with Transformer-Big architecture. Here we picked the \textit{Helsinki-NLP/opus-mt-ar-en} checkpoint. For tokenization, we instantiate our tokenizer which is based on SentencePiece \cite{KudoR18} 
with the \textit{AutoTokenizer.from\_pretrained} method. This ensures that the tokenizer corresponds to the model architecture we want to use. 

\subsubsubsection{NLLB:}
No-Language-Left-Behind (NLLB) \cite{nllbteam2022language} is a multilingual model which supports 200 languages with a massive size Transformer. Fine-tuning is carried out on NLLB using its distilled version \textit{facebook/nllb-200-distilled-600M} checkpoint. For tokenization, we instantiate a multilingual model provided by NLLB for tokenization with the \textit{NllbTokenizerFast.from\_pretrained} method. This ensures that the tokenizer corresponds to the model architecture we are using.

\subsubsubsection{ChatGPT3.5:}
We use the chatGPT-3.5-turbo model via its official API \footnote{https://chat.openai.com} which power the ChatGPT. Here we prepare our dataset in the format that is acceptabted by the API functions. In particular we convert the AR-EN pairs into the Prompt template for sentence-level translation as recommended in the OpenAI playground for sentence-level translation task.  In order to avoid error, we truncate all the sentence pairs to a max size of 4,290 characters before sending the request. Moreover, we set the size of the total tokens to about 378,460 tokens due to limit rate costs. For this model, we formatted the requests with as the system message first \textit{'You are a professional translator in the financial domain. Translate the following Arabic sentence: } ar\_en \textit{into English'} followed by user content messages, where ar\_en represent the AR-EN pairs.

Before, starting with the experiments, we considered the following three setups for fine-tuning the models on the domain-specific dataset. Next section 3 will discuss the results and findings.
\subsubsubsection{Setup \#1: Baseline models}\\
We consider pre-trained NMT models evaluated on our cleaned authentic test split containing 1000 AR-EN sentence pairs. Our baseline NMT models use the OPUS(bt-big) \footnote{https://github.com/Helsinki-NLP/Tatoeba-Challenge/tree/master/models/ara-eng} \cite{Tiedemann2020OPUSMTB}, NLLB 600M \footnote{https://huggingface.co/facebook/nllb-200-distilled-600M} \cite{nllbteam2022language} and ChatGPT-3.5 \footnote{https://platform.openai.com/docs/guides/gpt/chat-completions-api}.
\subsubsubsection{Setup \#2: Fine-tuning with authentic data}\\
For fine-tuning, we have initialized the transformer models with the trained weights of the baselines. We use our authentic dataset with the splits shown in table \ref{tab:datasplit-table}. We have kept all hyperparameters identical. The models have been fine-tuned until convergence over the validation set. At test time, the respective testset from the authentic dataset is used for this setup as well. Again, all metrics are reported.
\subsubsubsection{Setup \#3: Fine-tuning with augmented data}\\
Similar to previous setup, we have initialized the transformer models with the trained weights of the baselines. However, here we use our authentic dataset augmented with the respective data with the splits shown in table \ref{tab:datasplit-table}. Basically, we augment the authentic dataset with back-translated data and shuffle it. The same applies for synthetic data. This step yield two version of fine-tuning, one using the former and one using the latter.
The models have been fine-tuned until convergence over the validation set. At test time, the testset from the authentic dataset is used for this setup as well while also reporting all metrics.

\subsection{Metrics}
As performance measures, we report the spBLEU score \cite{papineni2002bleu} which uses a SentencePiece tokenizer, chrF \cite{popovic-2015-chrf}, TER \cite{snover2006study}, which are implemented in sacrebleu \footnote{https://github.com/mjpost/sacreBLEU}. Additionally, we compute COMET \cite{rei-etal-2020-comet} that was proposed recently by taking advantage of cross-lingual pre-trained LMs using knowledge from both source and target languages. COMET make prediction score that correlates with human judgement \cite{rei-etal-2020-comet}. For our experiments, we adapt the official COMET implementation \footnote{https://github.com/Unbabel/COMET}. For COMET, we use the reference-based Estimation model \textit{wmt20-comet-da}, trained based on (Direct Assessment) DA and used Quality Estimation (QE). Another score that correlates with human evaluation BERTScore 
  \cite{zhang2020bertscore} is also computed. Including different metrics in the evaluation allows us to test the models on metrics different from those used for training.
\section{Results and Discussion}
This section elaborates on our automatic and human evaluations and discusses the results. We also provide a preliminary comparison of the models' performance on domain-specific MT as baseline models and as fine-tuned models. Therefore, we report if they can perform robustly well on domain-specific or even noisy sentences from our collected dataset. Specifically, we focus on the translation robustness of the models on the translation of Arabic financial news.
Table \ref{tab:my-table} shows the main results over the respective testset. The ↑ and ↓ symbols in the tables indicate which values are better. We analyze the translation outputs by comparing the MT evaluation metrics in each setup.

\subsection{Automatic evaluation}
In setup \#1, OPUS and NLLB perform equally with inferior performances of around 14 and 42 for BLEU and chrF points, respectively. The TER score which is expressed as the ratio of the number of edits to the average number of words in the reference is high for the two models. Thus it indicates that the translation is of poor quality.
In terms of COMET score, both models have very poor results which means reference-based COMET may lose information from source, translation output, or reference embeddings, except for ChatGPT-3.5. But, BERTScores for all three models are high which means that they don't correlate with COMET score. In comparison, BERTScore and COMET have a  significant difference in their scores.
In contrast, ChatGPT-3.5 performs competitively better (BLEU 26.13) than OPUS and NLLB models. Indeed, we are not surprised by this fact which is in-line with related research works \cite{Peng2023TowardsMT, khondaker2023gptaraeval, FatenKhoshafah}. However, these findings are not consistent with a previous finding  \cite{alyafeai2023taqyim} where the authors evaluated ChatGPT and GPT on 4,000 Arabic-English pairs and found out that SoTA models like araT5 \cite{nagoudi2022arat5} outperforms ChatGPT by 19 BLEU Points.
As for the translation robustness, results from this setup\#1  suggest that ChatGPT-3.5 outperforms these models on financial news by a significant margin. \\

When, we analyze setup \#2, as expected, fine-tuning all models on authentic data has generally helped improve the BLEU scores and other metrics as well. This finding is also in-line with others' research \cite{KudoR18, moslem2022domain}. However, \cite{Peng2023TowardsMT} notice that for domain-specific translation (e.g., in the biomedical filed), ChatGPT’s performance degrades considerably. We attribute this behaviour to the observation that ChatGPT is capable to translate our sentences better than terminologies in sentences from the biomedical domain, a very specific domain. Furthermore, we clearly, see that BLEU scores increase from 14.58 to 48.83, from 14.38 to 43.43 and from 26.13 to 51.15 for OPUS, NLLB and chatGPT-3.5, respectively. In terms of COMET and BERTScore, both metrics correlate. This indicates acceptable translation outputs.\\
Concerning ChatGPT, even though it only used 2000 pairs of AR-EN sentences for fine-tuning, it outperforms all other models which means the MT quality of chatGPT can easily be improved with little additional data from the language pair, a fact that has not been previously confirmed for related approaches, since this is the first work that assesses the performance of ChatGPT fine-tuned models for AR-EN MT task. Nevertheless, for English, this work \cite{hendy2023good} has shown that ChatGPT has great robust translation capabilities over related SoTA MT models. Our experimental result confirms the latter finding and shows that with a carefully prepared certain amount of fine-tuning data, this model is capable to create acceptable translations. As for the translation robustness, results from this setup\#2  suggest that ChatGPT-3.5 performs competitively well on financial news. Regarding the human evaluation, all models in this setup reached possible and acceptable translations. We conclude, our experiment shows that providing in-domain examples to ChatGPT achieves comparable results to a SoTA model in terms of automatic and human evaluation.\\

In setup \#3 we fine-tune the baseline models with the augmented data in two versions, one using back-translated data and the other using synthetic data. We observe both lexical metrics (BLEU and chrF) show consistent degradation with all models. The same applies to TER score. 
For instance, for ChatGPT, BLEU score decreased dramatically from 51.15 to 34.67 when fine-tuned on synthetic data while maintaining an acceptable score (BLEU 45.38) when fine-tuned on back-translated data. We observe that COMET score degraded massively for ChatGPT more than for OPUS and NLLB. One explanation could be that the synthetic data may have a lot of generated tokens that are grammatically correct, but they have nonsense meaning, as we know from the current state of the generative text. This could indicate that the translation results are not close in embedding space with the source and reference. In contrast, BERTScore maintained a good score over the two versions for all models. 
In this setup, OPUS (bt-big) FT (back MT) has made it the best model that provides reasonably good scores translations, however still lags behind the OPUS model fine-tuned on authentic data by at least 1.3 BLEU points. 

Generally, the drop in performance for all models in this setup is not consistent with others' research. For instance, the authors in \cite{moslem2022domain} used synthetic data in the healthcare domain and achieved improvements on the in-domain test set. In comparison, with this work, the authors applied synthetic data generation using mGPT \footnote{https://huggingface.co/sberbank-ai/mGPT} a multilingual language model. We argue that this model might have better perplexity in generated tokens compared to araGPT2. To the best of our knowledge, we did not find any research work investigating the performance of both models in regard to Arabic. We will further investigate this issue in future work. However, there are many general reasons explaining OPUS, NLLB and ChatGPT behaviour in domain-specific MT, especially in the case of augmenting the dataset with synthetic data. One explanation is that the use of synthetic data may cause incorrect token choices, grammatical errors, or unnatural sentence structures to propagate into the translation outputs which make suboptimal translation outputs.

Indeed, the results of this study demonstrate the models' robust translation capabilities for in-domain adaptation. They perform well when fine-tuned on authentic data. However, we observe a discrepancy between COMET and BERTScore. For instance, ChatGPT-3.5 perform worse on augmented data yielding a lower COMET score (23.03) but still having high BERTScore (0.91). This behaviour seems uncommon. One possible explanation is that COMET with reference-based translation is failing to find closeness in all three resource embeddings, whereas BERTScore is able the find closeness in the similarity between an MT output and a reference translation.
This behaviour encourages us to drive human evaluation a much-needed score for trustworthiness.

\begin{figure}[htbp]
\centering
\begin{subfigure}{0.5\textwidth}
    \centering
    \includegraphics[width=\textwidth]{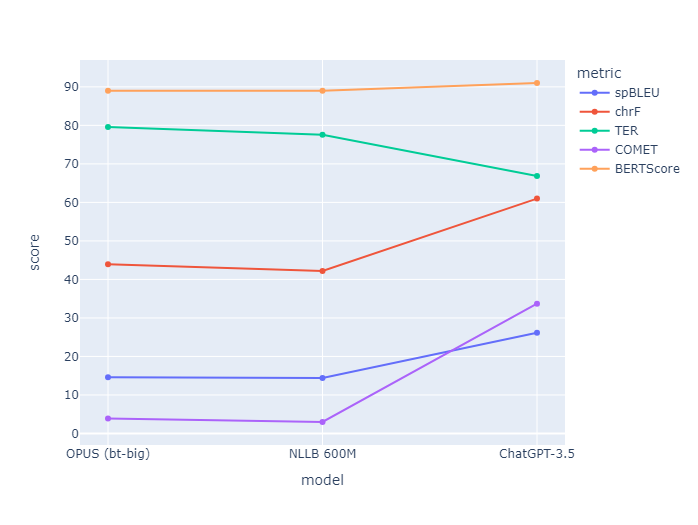}
    \caption{Setup \#1}
    \label{fig:setup1}
\end{subfigure}
     
\begin{subfigure}{0.5\textwidth}
    \centering
    \includegraphics[width=\textwidth]{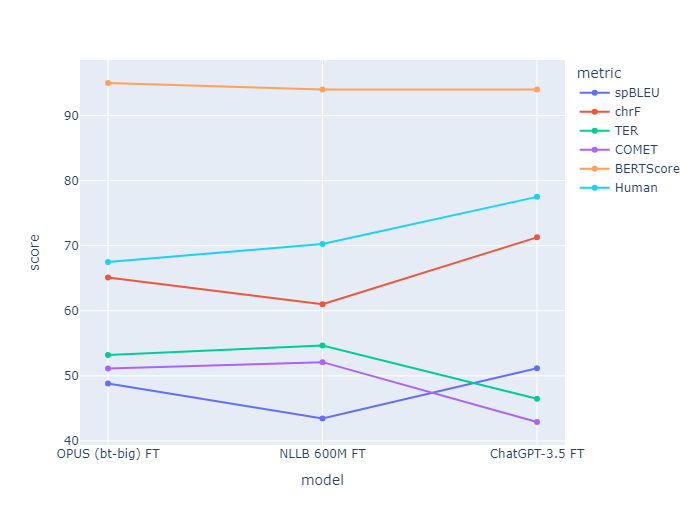}
    \caption{Setup \#2}
    \label{fig:setup2}
\end{subfigure}
 \hfill
\begin{subfigure}{0.5\textwidth}
    \centering
    \includegraphics[width=\textwidth]{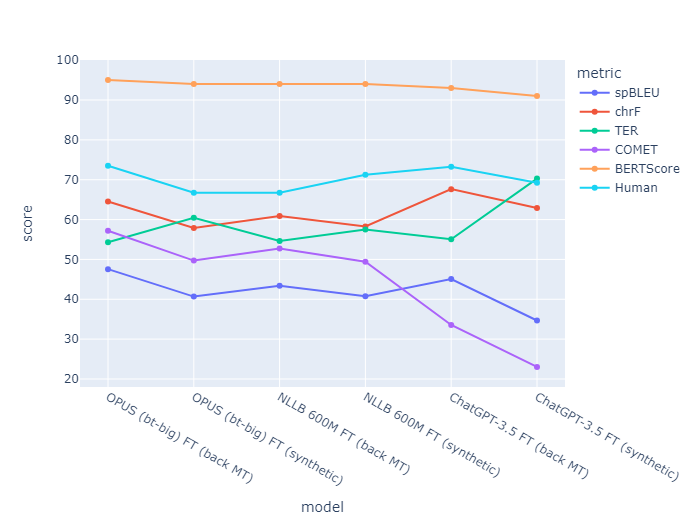}
    \caption{Setup \#3}
    \label{fig:setup3}
\end{subfigure}
 \hfill
\caption{Plotting the performance of the three models across different setups}
\label{fig:plots}
\centering
\end{figure}

Regarding the human scoring, we observe that in setup \#1 and \#2, where ChatGPT-3.5 made it the best model in terms of lexical and semantic metrics, the human evaluation supports this result with the highest score of 3.1. 
\\
However, in setup \#3, even though the automatic metrics are degraded for all models, except BERTScore, the human evaluation shows that the translation quality of all models is comparable.
Thus, we find that BERTScore correlates with human judgment more than COMET which has become recently a new state-of-the-art level of correlation with human judgment. This finding opens a great investigation for the future into whether semantic metrics correlate with human judgment and to what extent, in particular when ChatGPT-3.5 is applied. Figure \ref{fig:plots} elaborates on all the automatic evaluation and human results.


\begin{table}[]
\centering
\caption{MT evaluation scores and human evaluation for AR-EN Test dataset (1000 pairs). The best scores are in \textbf{bold}.}
\begin{tabular}{@{}llllllll@{}}
\toprule
\textbf{Setups} &
  \textbf{Model} &
  \textbf{spBLEU ↑} &
  \textbf{chrF ↑} &
  \textbf{TER ↓} &
  \textbf{COMET ↑} &
  \textbf{BERTScore ↑} &
    \textbf{Human ↑}\\ \midrule
\multirow{4}{*}{1} &
  OPUS (bt-big \tablefootnote{https://huggingface.co/Helsinki-NLP/opus-mt-ar-en}) &
 14.58 &
  43.93&
  79.59&
   3.89&
 0.89&
 -\\
 &
  NLLB 600M \tablefootnote{https://huggingface.co/facebook/nllb-200-distilled-600M} &
  14.38 &
  42.17 &
  77.58 &
  2.98&
  0.89&
  -\\
 &
  ChatGPT-3.5 &
  \textbf{26.13} &
  \textbf{60.98} &
  \textbf{66.83} &
  \textbf{33.7} &
  \textbf{0.91}&
  -\\
  \midrule
\multirow{4}{*}{2} & OPUS (bt-big) FT \tablefootnote{https://huggingface.co/asas-ai/opus-mt-ar-en-finetuned-ar-to-en}&
48.83&
65.11&
53.18 &
51.12 &
\textbf{0.95} &
 2.7 \\  
 & 
 NLLB 600M  FT \tablefootnote{https://huggingface.co/asas-ai/nllb-200-distilled-600M-finetuned-ar-to-en}   &
 43.43 &
 61.01 &
 54.65 &
 \textbf{52.10} &
 0.94 &
 2.81\\
 
 &
 ChatGPT-3.5 FT &
 \textbf{51.15} &
 \textbf{71.28} & 
 \textbf{46.47} & 
 42.90 & 
 0.94 &
 3.1\\
 \midrule
\multirow{8}{*}{3} 
& OPUS (bt-big) FT (back MT)\tablefootnote{https://huggingface.co/asas-ai/opus-mt-ar-en-finetuned\_augmented\_MT-ar-to-en} &    
      
\textbf{47.56}    &
64.53 & 
\textbf{54.30} &  
\textbf{57.21} & 
\textbf{0.95} &
2.94 \\
      
&
OPUS (bt-big) FT (synthetic) \tablefootnote{https://huggingface.co/asas-ai/opus-mt-ar-en-finetuned\_augmented\_synthetic-ar-to-en} &
      
40.67  & 
57.87  &
60.46 &  
49.71 & 
0.94  &
2.67 \\\cline{2-8} \\
& NLLB 600M FT (back MT)\tablefootnote{https://huggingface.co/asas-ai/nllb-200-distilled-600M-finetuned\_augmented\_MT\_ar-to-en} 
      &  
43.38& 
60.92& 
54.63 & 
52.77 & 
0.94 &
2.67\\
            
      &  
NLLB 600M FT (synthetic) \tablefootnote{https://huggingface.co/asas-ai/nllb-200-distilled-600M-finetuned\_augmented\_synthetic\_ar-to-en}     
     & 
     
40.77   & 
58.26  &  
57.48 &
49.44 & 
0.94     &
2.85 \\\cline{2-8} \\
     
& ChatGPT-3.5 FT (back MT)  
    &    
45.07  & 
\textbf{67.64}  & 
55.07 & 
33.55&
0.93 & 
2.93 \\

& ChatGPT-3.5 FT (synthetic)  &
34.67    &
62.93  &
70.29   &
23.03  &
0.91 &
 2.77     \\                      
     &  &  &  &  &
                          \\  
                          
                          \bottomrule
                          &

   FT = Fine-tuned  
   &  &  &  & 
\end{tabular}

\label{tab:my-table}
\end{table}

\subsection{Human evaluation}
In addition to the automatic evaluations reported above, we decided to assess the quality of our models' translations using human evaluation. To this end, we recruited three native speakers and domain experts (post-graduate students in finance) to rate the acceptability of 50 randomly selected sentences from the test set. Similar to \cite{moslem2022domain}, we conducted a bilingual evaluation, whereby the evaluators rated both the original source sentences and translations generated by the MT models. The human evaluators were asked to rate each of the sentence based on the scale proposed by \cite{coughlin2003correlating}, ranging from 1 to 4, and outlined as follows: 

\begin{itemize}
    \item 4 = Ideal: Not necessarily a perfect translation, but grammatically correct, with all information
accurately transferred.
    \item  3 = Acceptable: Not perfect (stylistically or grammatically odd), but definitely comprehensible,
    AND with accurate transfer of all important information.
    \item 2 = Possibly Acceptable: Possibly comprehensible (given enough context and/or time to work it
out); some information transferred accurately.
    \item 1 = Unacceptable: Absolutely not comprehensible and/or little or no information is accurately
transferred.
\end{itemize}

We first asked the three human evaluators to rate one model's output and then we conducted an inter-rater reliability analysis on their ratings. The result of weighted Cohen's Kappa is  X. Then, we asked each rater rate there models' outputs and provide justification for their responses were "Ideal" or "Unacceptable." The mean value of the raters' scores was averaged for each system, as shown in table \ref{tab:my-table}. 



\section{Conclusion}
In this paper, we conducted several experiments to
assess the performance of pre-trained NMT and LLM like GPT-3.5 using data augmentation in the domain of Arabic financial news articles. Generally, the results obtained from these experiments
are very promising. While ChatGPT shows good
results using few pairs, other models need more examples and still have lower performance. 
We explored the effectiveness of all models using data augmentation in the financial domain and found that the MT quality decreased for all models adequately. Here ChatGPT shows inferior performance, while OPUS still performs well on back-translated data than on synthetic data.

There are many future works that can be carried out based on the findings from this study. Firstly, we would like to explore new techniques and methods to enhance translation outputs rather than the approach of data augmentation. Secondly, we think it is valuable
to integrate more high-performance automatic
metrics into the comparison that take semantics into consideration in a better way than in COMET and BERTScore. 
Finally, we will explore novel approaches to integrate additional models or even incorporate domain-specific models for improved translation performance. 


\subsection{Acknowledgements}

This work is supported by a research grant from the Saudi Ministry of Culture. 

\printbibliography

@article{popel2020transforming,
  title={Transforming machine translation: a deep learning system reaches news translation quality comparable to human professionals},
  author={Popel, Martin and Tomkova, Marketa and Tomek, Jakub and Kaiser, {\L}ukasz and Uszkoreit, Jakob and Bojar, Ond{\v{r}}ej and {\v{Z}}abokrtsk{\`y}, Zden{\v{e}}k},
  journal={Nature communications},
  volume={11},
  number={1},
  pages={1--15},
  year={2020},
  publisher={Nature Publishing Group}
}

@article{bapna2019simple,
  title={Simple, scalable adaptation for neural machine translation},
  author={Bapna, Ankur and Arivazhagan, Naveen and Firat, Orhan},
  journal={arXiv preprint arXiv:1909.08478},
  year={2019}
}

@article{moslem2022domain,
  title={Domain-Specific Text Generation for Machine Translation},
  author={Moslem, Yasmin and Haque, Rejwanul and Kelleher, John D and Way, Andy},
  journal={arXiv preprint arXiv:2208.05909},
  year={2022}
}

@article{koehn2017six,
  title={Six challenges for neural machine translation},
  author={Koehn, Philipp and Knowles, Rebecca},
  journal={arXiv preprint arXiv:1706.03872},
  year={2017}
}

@incollection{hatem2010syntactic,
  title={Syntactic reordering for Arabic-English phrase-based machine translation},
  author={Hatem, Arwa and Omar, Nazlia},
  booktitle={Database theory and application, bio-science and bio-technology},
  pages={198--206},
  year={2010},
  publisher={Springer}
}

@article{almahasees2018assessment,
  title={Assessment of Google and Microsoft Bing translation of journalistic texts},
  author={Almahasees, Zakaryia Mustafa},
  journal={International Journal of Languages, Literature and Linguistics},
  volume={4},
  number={3},
  pages={231--235},
  year={2018}
}

@article{sennrich2015neural,
  title={Neural machine translation of rare words with subword units},
  author={Sennrich, Rico and Haddow, Barry and Birch, Alexandra},
  journal={arXiv preprint arXiv:1508.07909},
  year={2015}
}

@article{costa2016character,
  title={Character-based neural machine translation},
  author={Costa-Jussa, Marta R and Fonollosa, Jos{\'e} AR},
  journal={arXiv preprint arXiv:1603.00810},
  year={2016}
}

@article{luong2016achieving,
  title={Achieving open vocabulary neural machine translation with hybrid word-character models},
  author={Luong, Minh-Thang and Manning, Christopher D},
  journal={arXiv preprint arXiv:1604.00788},
  year={2016}
}

@article{muller2019domain,
  title={Domain robustness in neural machine translation},
  author={M{\"u}ller, Mathias and Rios, Annette and Sennrich, Rico},
  journal={arXiv preprint arXiv:1911.03109},
  year={2019}
}

@article{oudah2019impact,
  title={The impact of preprocessing on Arabic-English statistical and neural machine translation},
  author={Oudah, Mai and Almahairi, Amjad and Habash, Nizar},
  journal={arXiv preprint arXiv:1906.11751},
  year={2019}
}

@article{yang2020survey,
  title={A survey of deep learning techniques for neural machine translation},
  author={Yang, Shuoheng and Wang, Yuxin and Chu, Xiaowen},
  journal={arXiv preprint arXiv:2002.07526},
  year={2020}
}

@article{sutskever2014sequence,
  title={Sequence to sequence learning with neural networks},
  author={Sutskever, Ilya and Vinyals, Oriol and Le, Quoc V},
  journal={Advances in neural information processing systems},
  volume={27},
  year={2014}
}

@article{johnson2017google,
  title={Google’s multilingual neural machine translation system: Enabling zero-shot translation},
  author={Johnson, Melvin and Schuster, Mike and Le, Quoc V and Krikun, Maxim and Wu, Yonghui and Chen, Zhifeng and Thorat, Nikhil and Vi{\'e}gas, Fernanda and Wattenberg, Martin and Corrado, Greg and others},
  journal={Transactions of the Association for Computational Linguistics},
  volume={5},
  pages={339--351},
  year={2017},
  publisher={MIT Press}
}

@article{bahdanau2014neural,
  title={Neural machine translation by jointly learning to align and translate},
  author={Bahdanau, Dzmitry and Cho, Kyunghyun and Bengio, Yoshua},
  journal={arXiv preprint arXiv:1409.0473},
  year={2014}
}

@article{luong2015effective,
  title={Effective approaches to attention-based neural machine translation},
  author={Luong, Minh-Thang and Pham, Hieu and Manning, Christopher D},
  journal={arXiv preprint arXiv:1508.04025},
  year={2015}
}

@article{zhou2016deep,
  title={Deep recurrent models with fast-forward connections for neural machine translation},
  author={Zhou, Jie and Cao, Ying and Wang, Xuguang and Li, Peng and Xu, Wei},
  journal={Transactions of the Association for Computational Linguistics},
  volume={4},
  pages={371--383},
  year={2016},
  publisher={MIT Press}
}

@article{ataman2019latent,
  title={A latent morphology model for open-vocabulary neural machine translation},
  author={Ataman, Duygu and Aziz, Wilker and Birch, Alexandra},
  journal={arXiv preprint arXiv:1910.13890},
  year={2019}
}

@article{meng2015encoding,
  title={Encoding source language with convolutional neural network for machine translation},
  author={Meng, Fandong and Lu, Zhengdong and Wang, Mingxuan and Li, Hang and Jiang, Wenbin and Liu, Qun},
  journal={arXiv preprint arXiv:1503.01838},
  year={2015}
}

@article{gehring2016convolutional,
  title={A convolutional encoder model for neural machine translation},
  author={Gehring, Jonas and Auli, Michael and Grangier, David and Dauphin, Yann N},
  journal={arXiv preprint arXiv:1611.02344},
  year={2016}
}

@article{kalchbrenner2016neural,
  title={Neural machine translation in linear time},
  author={Kalchbrenner, Nal and Espeholt, Lasse and Simonyan, Karen and Oord, Aaron van den and Graves, Alex and Kavukcuoglu, Koray},
  journal={arXiv preprint arXiv:1610.10099},
  year={2016}
}

@inproceedings{gehring2017convolutional,
  title={Convolutional sequence to sequence learning},
  author={Gehring, Jonas and Auli, Michael and Grangier, David and Yarats, Denis and Dauphin, Yann N},
  booktitle={International conference on machine learning},
  pages={1243--1252},
  year={2017},
  organization={PMLR}
}

@article{vaswani2017attention,
  title={Attention is all you need},
  author={Vaswani, Ashish and Shazeer, Noam and Parmar, Niki and Uszkoreit, Jakob and Jones, Llion and Gomez, Aidan N and Kaiser, {\L}ukasz and Polosukhin, Illia},
  journal={Advances in neural information processing systems},
  volume={30},
  year={2017}
}

@article{stahlberg2020neural,
  title={Neural machine translation: A review},
  author={Stahlberg, Felix},
  journal={Journal of Artificial Intelligence Research},
  volume={69},
  pages={343--418},
  year={2020}
}

@article{devlin2018bert,
  title={Bert: Pre-training of deep bidirectional transformers for language understanding},
  author={Devlin, Jacob and Chang, Ming-Wei and Lee, Kenton and Toutanova, Kristina},
  journal={arXiv preprint arXiv:1810.04805},
  year={2018}
}

@article{antoun2020arabert,
  title={Arabert: Transformer-based model for Arabic language understanding},
  author={Antoun, Wissam and Baly, Fady and Hajj, Hazem},
  journal={arXiv preprint arXiv:2003.00104},
  year={2020}
}

@inproceedings{lan2020gigabert,
  title={Gigabert: Zero-shot transfer learning from english to arabic},
  author={Lan, Wuwei and Chen, Yang and Xu, Wei and Ritter, Alan},
  booktitle={Proceedings of The 2020 Conference on Empirical Methods on Natural Language Processing (EMNLP)},
  year={2020}
}

@inproceedings{deng2020factorized,
  title={Factorized transformer for multi-domain neural machine translation},
  author={Deng, Yongchao and Yu, Hongfei and Yu, Heng and Duan, Xiangyu and Luo, Weihua},
  booktitle={Findings of the Association for Computational Linguistics: EMNLP 2020},
  pages={4221--4230},
  year={2020}
}

@article{ehab2019english,
  title={English-Arabic hybrid machine translation system using EBMT and translation memory},
  author={Ehab, Rana and Amer, Eslam and Gadallah, Mahmoud},
  journal={Int. J. Adv. Comput. Sci. Appl.},
  volume={10},
  number={1},
  pages={195--203},
  year={2019}
}

@inproceedings{abdul2019exploring,
  title={Exploring transfer learning and domain data selection for the biomedical translation},
  author={Abdul-Rauf, Sadaf and Kiani, Kiran and Zafar, Ammara and Nawaz, Raheel and others},
  booktitle={Proceedings of the Fourth Conference on Machine Translation (Volume 3: Shared Task Papers, Day 2)},
  pages={156--163},
  year={2019}
}

@article{chu2020survey,
  title={A survey of domain adaptation for machine translation},
  author={Chu, Chenhui and Wang, Rui},
  journal={Journal of information processing},
  volume={28},
  pages={413--426},
  year={2020},
  publisher={Information Processing Society of Japan}
}

@inproceedings{daume2011domain,
  title={Domain adaptation for machine translation by mining unseen words},
  author={Daum{\'e} Iii, Hal and Jagarlamudi, Jagadeesh},
  booktitle={Proceedings of the 49th Annual Meeting of the Association for Computational Linguistics: Human Language Technologies},
  pages={407--412},
  year={2011}
}

@article{ameur2020arabic,
  title={Arabic machine translation: A survey of the latest trends and challenges},
  author={Ameur, Mohamed Seghir Hadj and Meziane, Farid and Guessoum, Ahmed},
  journal={Computer Science Review},
  volume={38},
  pages={100305},
  year={2020},
  publisher={Elsevier}
}

@article{zakraoui2021arabic,
  title={Arabic Machine Translation: A Survey With Challenges and Future Directions},
  author={Zakraoui, Jezia and Saleh, Moutaz and Al-Maadeed, Somaya and Alja’am, Jihad Mohamed},
  journal={IEEE Access},
  volume={9},
  pages={161445--161468},
  year={2021},
  publisher={IEEE}
}

@article{habash2010introduction,
  title={Introduction to Arabic natural language processing},
  author={Habash, Nizar Y},
  journal={Synthesis lectures on human language technologies},
  volume={3},
  number={1},
  pages={1--187},
  year={2010},
  publisher={Morgan \& Claypool Publishers}
}

@inproceedings{antoun-etal-2021-aragpt2,
    title = "{A}ra{GPT}2: Pre-Trained Transformer for {A}rabic Language Generation",
    author = "Antoun, Wissam  and
      Baly, Fady  and
      Hajj, Hazem",
    booktitle = "Proceedings of the Sixth Arabic Natural Language Processing Workshop",
    month = apr,
    year = "2021",
    address = "Kyiv, Ukraine (Virtual)",
    publisher = "Association for Computational Linguistics",
    url = "https://aclanthology.org/2021.wanlp-1.21",
    pages = "196--207"
}

@inproceedings{Tiedemann2020OPUSMTB,
  title={OPUS-MT – Building open translation services for the World},
  author={J{\"o}rg Tiedemann and Santhosh Thottingal},
  booktitle={European Association for Machine Translation Conferences/Workshops},
  year={2020},
  url={https://api.semanticscholar.org/CorpusID:221097277}
}

@misc{nllbteam2022language,
      title={No Language Left Behind: Scaling Human-Centered Machine Translation}, 
      author={NLLB Team and Marta R. Costa-jussà and James Cross and Onur Çelebi and Maha Elbayad and Kenneth Heafield and Kevin Heffernan and Elahe Kalbassi and Janice Lam and Daniel Licht and Jean Maillard and Anna Sun and Skyler Wang and Guillaume Wenzek and Al Youngblood and Bapi Akula and Loic Barrault and Gabriel Mejia Gonzalez and Prangthip Hansanti and John Hoffman and Semarley Jarrett and Kaushik Ram Sadagopan and Dirk Rowe and Shannon Spruit and Chau Tran and Pierre Andrews and Necip Fazil Ayan and Shruti Bhosale and Sergey Edunov and Angela Fan and Cynthia Gao and Vedanuj Goswami and Francisco Guzmán and Philipp Koehn and Alexandre Mourachko and Christophe Ropers and Safiyyah Saleem and Holger Schwenk and Jeff Wang},
      year={2022},
      eprint={2207.04672},
      archivePrefix={arXiv},
      primaryClass={cs.CL}
}

@misc{nagoudi2022arat5,
      title={AraT5: Text-to-Text Transformers for Arabic Language Generation}, 
      author={El Moatez Billah Nagoudi and AbdelRahim Elmadany and Muhammad Abdul-Mageed},
      year={2022},
      eprint={2109.12068},
      archivePrefix={arXiv},
      primaryClass={cs.CL}
}

@misc{khondaker2023gptaraeval,
      title={GPTAraEval: A Comprehensive Evaluation of ChatGPT on Arabic NLP}, 
      author={Md Tawkat Islam Khondaker and Abdul Waheed and El Moatez Billah Nagoudi and Muhammad Abdul-Mageed},
      year={2023},
      eprint={2305.14976},
      archivePrefix={arXiv},
      primaryClass={cs.CL}
}

@misc{FatenKhoshafah,
  author={Khoshafah, Faten },
  journal={Research Square},
  title ={ChatGPT for Arabic-English Translation: Evaluating the Accuracy},
  pages={https://doi.org/10.21203/rs.3.rs-2814154/v1},
  year={2023},
  Publisher = {Research Square}

}

@article{alyafeai2023taqyim,
  title={Taqyim: Evaluating Arabic NLP Tasks Using ChatGPT Models},
  author={Alyafeai, Zaid and Alshaibani, Maged S and AlKhamissi, Badr and Luqman, Hamzah and Alareqi, Ebrahim and Fadel, Ali},
  journal={arXiv preprint arXiv:2306.16322},
  year={2023}
}

@misc{hendy2023good,
      title={How Good Are GPT Models at Machine Translation? A Comprehensive Evaluation}, 
      author={Amr Hendy and Mohamed Abdelrehim and Amr Sharaf and Vikas Raunak and Mohamed Gabr and Hitokazu Matsushita and Young Jin Kim and Mohamed Afify and Hany Hassan Awadalla},
      year={2023},
      eprint={2302.09210},
      archivePrefix={arXiv},
      primaryClass={cs.CL}
}

@inproceedings{papineni2002bleu,
  author = {Papineni, Kishore and Roukos, Salim and Ward, Todd and Zhu, Wei-Jing},
  booktitle = {Proceedings of the 40th annual meeting on association for computational linguistics},
  organization = {Association for Computational Linguistics},
  pages = {311--318},
  title = {BLEU: a method for automatic evaluation of machine translation},
  year = 2002
}

@misc{zhang2020bertscore,
      title={BERTScore: Evaluating Text Generation with BERT}, 
      author={Tianyi Zhang and Varsha Kishore and Felix Wu and Kilian Q. Weinberger and Yoav Artzi},
      year={2020},
      eprint={1904.09675},
      archivePrefix={arXiv},
      primaryClass={cs.CL}
}

@inproceedings{rei-etal-2020-comet,
    title = "{COMET}: A Neural Framework for {MT} Evaluation",
    author = "Rei, Ricardo  and
      Stewart, Craig  and
      Farinha, Ana C  and
      Lavie, Alon",
    booktitle = "Proceedings of the 2020 Conference on Empirical Methods in Natural Language Processing (EMNLP)",
    month = nov,
    year = "2020",
    address = "Online",
    publisher = "Association for Computational Linguistics",
    url = "https://aclanthology.org/2020.emnlp-main.213",
    doi = "10.18653/v1/2020.emnlp-main.213",
    pages = "2685--2702"
}

@inproceedings{popovic-2015-chrf,
    title = "chr{F}: character n-gram {F}-score for automatic {MT} evaluation",
    author = "Popovi{\'c}, Maja",
    booktitle = "Proceedings of the Tenth Workshop on Statistical Machine Translation",
    month = sep,
    year = "2015",
    address = "Lisbon, Portugal",
    publisher = "Association for Computational Linguistics",
    url = "https://aclanthology.org/W15-3049",
    doi = "10.18653/v1/W15-3049",
    pages = "392--395"
}

@inproceedings{snover2006study,
  title={A study of translation edit rate with targeted human annotation},
  author={Snover, Matthew and Dorr, Bonnie and Schwartz, Richard and Micciulla, Linnea and Makhoul, John},
  booktitle={Proceedings of the 7th Conference of the Association for Machine Translation in the Americas: Technical Papers},
  pages={223--231},
  year={2006}
}

@inproceedings{coughlin2003correlating,
  title={Correlating automated and human assessments of machine translation quality},
  author={Coughlin, Deborah},
  booktitle={Proceedings of Machine Translation Summit IX: Papers},
  year={2003}
}

@article{saunders2022domain,
  title={Domain adaptation and multi-domain adaptation for neural machine translation: A survey},
  author={Saunders, Danielle},
  journal={Journal of Artificial Intelligence Research},
  volume={75},
  pages={351--424},
  year={2022}
}

@article{radford2019language,
  title={Language Models are Unsupervised Multitask Learners},
  author={Radford, Alec and Wu, Jeff and Child, Rewon and Luan, David and Amodei, Dario and Sutskever, Ilya},
  year={2019}
}

@misc{poncelas1804investigating,
  title={Investigating Backtranslation in Neural Machine Translation. 2018},
  author={Poncelas, Alberto and Shterionov, Dimitar and Way, Andy and Wenniger, G and Passban, Peyman},
  year={1804},
  publisher={Arxiv}
}

@inproceedings{martinez2020customized,
  title={Customized Neural Machine Translation Systems for the Swiss Legal Domain},
  author={Mart{\'\i}nez-Dom{\'\i}nguez, Rub{\'e}n and Rikters, Mat{\=\i}ss and Vasi{\c{l}}evskis, Art{\=u}rs and Pinnis, M{\=a}rcis and Reichenberg, Paula},
  booktitle={Proceedings of the 14th Conference of the Association for Machine Translation in the Americas (Volume 2: User Track)},
  pages={217--223},
  year={2020}
}

@article{grangier2021trade,
  title={The trade-offs of domain adaptation for neural language models},
  author={Grangier, David and Iter, Dan},
  journal={arXiv preprint arXiv:2109.10274},
  year={2021}
}

@inproceedings{chu2017empirical,
  title={An empirical comparison of domain adaptation methods for neural machine translation},
  author={Chu, Chenhui and Dabre, Raj and Kurohashi, Sadao},
  booktitle={Proceedings of the 55th Annual Meeting of the Association for Computational Linguistics (Volume 2: Short Papers)},
  pages={385--391},
  year={2017}
}

@article{liu2021paramed,
  title={ParaMed: a parallel corpus for English--Chinese translation in the biomedical domain},
  author={Liu, Boxiang and Huang, Liang},
  journal={BMC Medical Informatics and Decision Making},
  volume={21},
  number={1},
  pages={258},
  year={2021},
  publisher={Springer}
}

@article{laubli2019post,
  title={Post-editing productivity with neural machine translation: An empirical assessment of speed and quality in the banking and finance domain},
  author={L{\"a}ubli, Samuel and Amrhein, Chantal and D{\"u}ggelin, Patrick and Gonzalez, Beatriz and Zwahlen, Alena and Volk, Martin},
  journal={arXiv preprint arXiv:1906.01685},
  year={2019}
}

@inproceedings{bakr2008hybrid,
  title={A hybrid approach for converting written Egyptian colloquial dialect into diacritized Arabic},
  author={Bakr, Hitham Abo and Shaalan, Khaled and Ziedan, Ibrahim},
  booktitle={The 6th international conference on informatics and systems, infos2008. cairo university},
  year={2008},
  organization={Citeseer}
}

@inproceedings{habash2009improving,
  title={Improving Arabic-Chinese statistical machine translation using English as pivot language},
  author={Habash, Nizar and Hu, Jun},
  booktitle={Proceedings of the Fourth Workshop on Statistical Machine Translation},
  pages={173--181},
  year={2009}
}

@inproceedings{mohamed2012transforming,
  title={Transforming standard Arabic to colloquial Arabic},
  author={Mohamed, Emad and Mohit, Behrang and Oflazer, Kemal},
  booktitle={Proceedings of the 50th Annual Meeting of the Association for Computational Linguistics (Volume 2: Short Papers)},
  pages={176--180},
  year={2012}
}

@article{nagoudi2021arat5,
  title={AraT5: Text-to-text transformers for Arabic language generation},
  author={Nagoudi, El Moatez Billah and Elmadany, AbdelRahim and Abdul-Mageed, Muhammad},
  journal={arXiv preprint arXiv:2109.12068},
  year={2021}
}

@inproceedings{tiedemann-2020-tatoeba,
    title = "The {T}atoeba {T}ranslation {C}hallenge {--} {R}ealistic Data Sets for Low Resource and Multilingual {MT}",
    author = {Tiedemann, J{\"o}rg},
    booktitle = "Proceedings of the Fifth Conference on Machine Translation",
    month = nov,
    year = "2020",
    address = "Online",
    publisher = "Association for Computational Linguistics",
    url = "https://www.aclweb.org/anthology/2020.wmt-1.139",
    pages = "1174--1182"
}

@inproceedings{KudoR18,
  author       = {Taku Kudo and
                  John Richardson},
  title        = {SentencePiece: {A} simple and language independent subword tokenizer
                  and detokenizer for Neural Text Processing},
  booktitle    = {Proceedings of the 2018 Conference on Empirical Methods in Natural
                  Language Processing, {EMNLP} 2018: System Demonstrations, Brussels,
                  Belgium, October 31 - November 4, 2018},
  pages        = {66--71},
  year         = {2018},
  url          = {https://doi.org/10.18653/v1/d18-2012},
  doi          = {10.18653/v1/d18-2012},

}

@article{moslem2023adaptive,
  title={Adaptive machine translation with large language models},
  author={Moslem, Yasmin and Haque, Rejwanul and Way, Andy},
  journal={arXiv preprint arXiv:2301.13294},
  year={2023}
}

@article{Peng2023TowardsMT,
  title={Towards Making the Most of ChatGPT for Machine Translation},
  author={Keqin Peng and Liang Ding and Qihuang Zhong and Li Shen and Xuebo Liu and Min Zhang and Yuanxin Ouyang and Dacheng Tao},
  journal={ArXiv},
  year={2023},
  volume={abs/2303.13780},
  url={https://api.semanticscholar.org/CorpusID:257704711}
}

\end{document}